\documentclass[journal]{IEEEtran}

\ifCLASSINFOpdf
\else
   \usepackage[dvips]{graphicx}
\fi
\usepackage{url}
\usepackage{makecell}
\usepackage{amsmath}
\usepackage{amssymb}
\usepackage{bbm}
\usepackage[numbers,sort&compress]{natbib}
\usepackage{hyperref}
\usepackage{graphicx}

\begin{document}

\title{JEP-KD: Joint-Embedding Predictive Architecture Based Knowledge Distillation for Visual Speech Recognition}

\author{Chang Sun, Hong Yang, and Bo Qin.
\thanks{
   Chang Sun is with the First Research Institute of the Ministry of Public Security of PRC and Beijing University of Posts and Telecommunications(e-mail: sunchang272@foxmail.com).
}
}

\markboth{Journal of \LaTeX\ Class Files, Vol. 14, No. 8, August 2015}
{Shell \MakeLowercase{\textit{et al.}}: Bare Demo of IEEEtran.cls for IEEE Journals}
\maketitle

\begin{abstract}
   Visual Speech Recognition (VSR) tasks are generally recognized to have a lower theoretical performance ceiling than Automatic Speech Recognition (ASR), 
   owing to the inherent limitations of conveying semantic information visually. 
   To mitigate this challenge, this paper introduces an advanced knowledge distillation approach using a Joint-Embedding Predictive Architecture (JEPA), named JEP-KD, 
   designed to more effectively utilize audio features during model training. 
   Central to JEP-KD is the inclusion of a generative network within the embedding layer, which enhances the video encoder's capacity for semantic feature extraction and brings it into closer alignment with the audio features from a pre-trained ASR model's encoder. 
   This approach aims to progressively reduce the performance gap between VSR and ASR. 
   Moreover, a comprehensive multimodal, multistage training regimen for the JEP-KD framework is established, bolstering the robustness and efficacy of the training process. 
   Experiment results demonstrate that JEP-KD significantly improves the performance of VSR models and demonstrates versatility across different VSR platforms, indicating its potential for broader application within other multimodal tasks.
\end{abstract}


\IEEEpeerreviewmaketitle

\section{Introduction}
\IEEEPARstart{V}{SR}, also known as lip-reading, 
is a domain of machine vision technology that translates sequences of lip movements from videos into corresponding text. 
Distinct from speech recognition, VSR is commonly utilized to facilitate communication in environments with significant background noise or where sound is absent \cite{lr-yingyong}. 
As the discipline advances, the video modality's inherent semantic information deficiency emerges as a prominent hurdle in VSR research, critically restricting the task's accuracy ceiling \cite{lr-wenti}. 
This issue is especially acute in Chinese lip-reading, where homophonic characters that share visual articulation patterns, as well as those with weak visual cues for lip-assisted pronunciation, predominate. 
The frequent occurrence of such characters within a sentence can greatly obscure the intended message, risking the failure of lip-reading systems.

\begin{figure}
\centerline{\includegraphics[width=0.85\columnwidth]{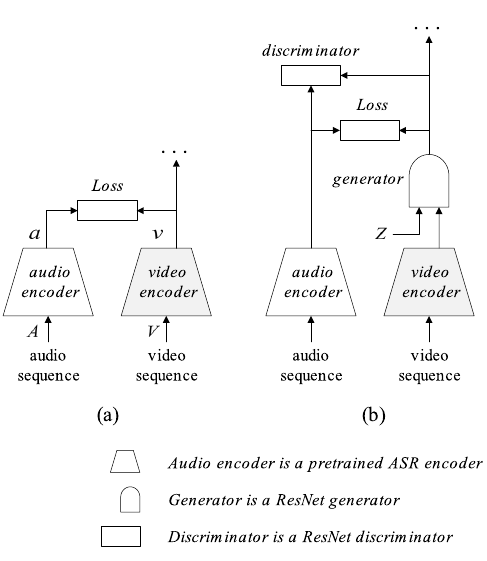}}
\caption{Comparison of the current knowledge distillation structure with the JEP-KD structure. (a) The knowledge distillation structure commonly used in current VSR tasks. (b) The JEP-KD structure proposed in this paper, which primarily differs from (a) in that JEP-KD introduces a prediction architecture at the embedding layer.}
\label{fig:fig1}
\end{figure}

Utilizing trained ASR models to conduct knowledge distillation \cite{zhishizhengliu} is a widely acknowledged and efficacious strategy to enhance performance. 
This process entails the integration of a pretrained ASR model to function as a teacher network, which during training, 
guides the VSR model by constraining its outputs and intermediary representations through the ASR model's analogous elements. 
Nonetheless, the prevalent method involves employing a loss function, such as L1 Loss, at the feature stage (refer to Figure \ref{fig:fig1} (a)) \cite{duibi1,duibi2,duibi3,duibi4}. 
Given the video modality's intrinsic semantic limitations, such rigid alignment tactics are challenging and typically yield suboptimal results. 
On one side, these methods inadequately exploit the auditory features, thus failing to facilitate a robust knowledge transfer; 
on the other end of the spectrum, studies have identified an intrinsic disparity across modalities akin to interlingual translation \cite{kuamotaiwenti}. 
We hold that the semantic gaps manifesting in video modalities relative to audio ones adhere to a systematic pattern — that is, 
these consistent omissions are correlated with sentence content and are, therefore, predictable. 
Inspired by the JEPA \cite{jepa1,jepa2}, introducing a tailored generative network into the semantic domain to augment the video semantic features could not only allow the video encoder to more foucsed on video feature extraction but also promote better alignment with the audio semantic features. 
Consequently, this enhances the VSR model's ability to mirror the performance of ASR models. 
Moreover, such a predictive framework can discern the semantic variances across modalities and construct a mapping interlinking the respective feature domains, 
thereby optimizing the interchange of knowledge.

It is widely recognized that the video modality encapsulates unique semantic information absent from the audio modality \cite{shipindayuyinpin}. 
So indiscriminately aligning video features with audio features may lead to the loss of such information. 
Nevertheless, considering the considerable performance gap that still exists between VSR and ASR, 
enhancing the capabilities of VSR as much as possible remains the most important research objective at the current stage. 
Additionally, employing an Audio-Visual Speech Recognition (AVSR) model as a guiding 'teacher network' presents a potential solution \cite{avsr1, avsr2, avsr3}. 
However, this paper focuses on discussing the feasibility of the proposed knowledge distillation structure. 
This module, as a plug-and-play enhancement structure, can be easily ported to other related tasks and play an effective role.

This paper's contributions are twofold: 
firstly, advancing a JEPA-based knowledge distillation architecture applicable to VSR and analogous tasks necessitating such frameworks; 
secondly, formulating a comprehensive training methodology comprising four models and three stages tailored to models invoking the JEP-KD architecture.

The structure of this paper is organized as follows: 
Section \ref{proposal} provides an in-depth presentation of the model architecture and delineates the multi-model, multi-stage training approach. 
Section \ref{experiment} explores the experimental setup and delineates the findings. 
The paper concludes with a summary and a list of references.

\begin{figure*}[h]
\centerline{\includegraphics[width=2\columnwidth]{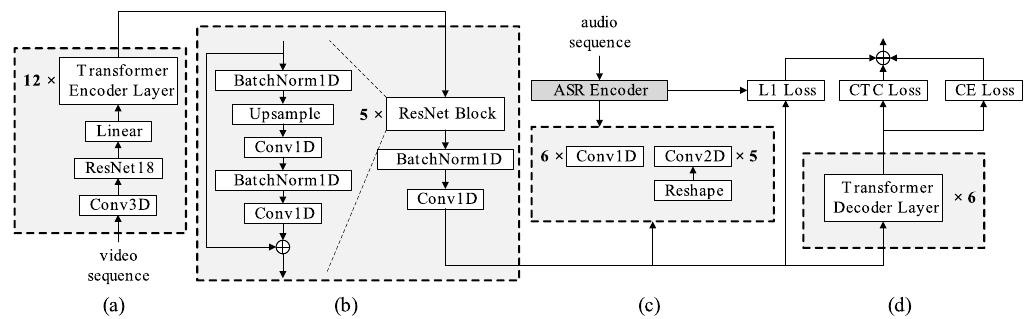}}
\caption{Visualization of the VSR model after the introduction of the JEP-KD structure. (a), (b), (c), and (d) are the structural diagrams of the video encoder, generator, discriminator, and decoder, respectively. During the training process, three types of loss functions—L1, CTC, and CE—are used for constraints.}
\label{fig:fig2}
\end{figure*}

\section{Proposal}
\label{proposal}
This section primarily discusses the fundamental principles and model architecture of JEP-KD, 
and provides a detailed introduction to the specific implementation processes utilized for validation of the structure, 
including the VSR model, ASR model, loss functions.

\subsection{Overview}
The principal architecture of JEP-KD is illustrated in Figure \ref{fig:fig1} (b). 
The commonly adopted knowledge distillation structure for the current VSR tasks involves adding an L1 loss or CE loss 
to constrain the output of the VSR encoder on the feature level so that it more closely approximates the output of the introduced pre-trained speech recognition model encoder.
As shown in Figure \ref{fig:fig1} (a), the audio encoder $f_{e A}(\cdot)$ takes audio sequence $X_A$ as input, where $a$ represents the output of audio encoder $f_{e A}\left(X_A ; \theta_A\right)$. 
The $a$ is a high-dimensional semantic vector that can be translated into text modality by audio decoder. Similarly, the $v$ is the output $f_{e V}\left(X_V ; \theta_V\right)$ of the video encoder, 
which is a vector that can be translated into text mode by video decoder. The usual knowledge distillation method adds a loss function $L$, usually L1 loss or CE loss, in the embedding layer. 
Generally speaking, during the training process of ASR tasks, 
the audio encoder gradually learns to remove the semantic-irrelevant information from the speech while retaining a vector $a$ carrying all the semantic information. 
Therefore, the vector $a$ is a subset of the speech signal that has been purified of semantic information. 
In other words, the semantic information is still carried by a high-dimensional representation of the speech signal.
Similarly, the output $v$ of the video encoder is also a semantic information representation based on the video signal. Even if the structure of the encoder is similar, $a$ and $v$ may still not be in the same dimension, 
but more like two different languages of the same semantics, with cross-modal differences. On the other hand, since lip movement is only an auxiliary movement for human vocalization, 
there are innate defects in semantic representation capabilities, resulting in incomplete semantics of the video modality. Therefore, the effect of this direct fitting constraint has certain limitations.

Therefore, we designed a joint embedding predictive knowledge distillation structure as shown in Figure \ref{fig:fig1} (b). 
Here we introduced a generator $G(\cdot)$, $z$ is a random additional variable, 
and $G\left(z,v\right)$ is used to eliminate the previously mentioned the modal differences between $a$ and $v$ in the embedding space and simulate the semantic loss of the video modality compared to the speech modality. 
Then we added a discriminator $D(\cdot)$ to supervise the training of $G\left(z,v\right)$ and $a$. And the similarity constraint $L$ between modalities is retained, 
but the input of $L$ is changed to $G\left(z,v\right)$ and $a$. 
In this structure, we reference the loss function of LS GAN \cite{lsgan} as shown in Equation \ref{lsgan_loss} to set the loss function for the predicted structure, 
where we set $a=c=1$, $b=0$, and incorporate $L\left(a,G\left(z,v\right)\right)$ into the loss function of the generator, 
with the complete loss function for $G(\cdot)$ and $D(\cdot)$ presented as shown in Equation \ref{jepkd_loss}.

{\small
\begin{equation}
\begin{aligned}
&\min _D J(D)=\min _D{ }_2^1 E_{x \sim P_r}[D(x)-a]^2+{ }_2^1 E_{z \sim P_z}[D(G(z))-b]^2\\
&\min _G J(G)=\min _G{ }_2^1 E_{z \sim P_z}[D(G(z))-c]^2
\end{aligned}
\label{lsgan_loss}
\end{equation}
}

{\small
\begin{equation}
\begin{aligned}
&\min _D J(D)=\min _D{ }_2^1 E_{x \sim P_r}[D(a)-1]^2+{ }_2^1 E_{z \sim P_z}D(G(z,v))^2\\
&\min _G J(G)=\min _G{ }_2^1 E_{z \sim P_z}[D(G(z,v))-1]^2+L(a, G(z, v))
\end{aligned}
\label{jepkd_loss}
\end{equation}
}

\subsection{Details}
In the practical implementation, 
we referred to Ma's work \cite{masmodel} and designed the model structure as shown in Figure \ref{fig:fig2}. 
To introduce the JEP-KD, we divided the recognition model into four stages as shown in Figure \ref{fig:fig2} (a)-(d). 
These are the encoder model, generator model, discriminator model, and decoder model. 
The encoder model consists of a frontend 3D convolution layer, a ResNet18 feature extraction layer, and 12 layers of Transformer encoder layer. 
The generator model primarily comprises 5 layers of ResNet2D Block. 
The discriminator is a hybrid model, consisting of a 1-dimensional discriminator made up of one 1-dimensional convolution and a 2-dimensional discriminator composed of one 2-dimensional convolution. 
The decoder is a standard 6-layer Transformer decoder layer. 

During the training process, we divide it into three stages. 
The first stage we call the warm-up phase, where we enable updates to the parameters of the encoder, 
generator, and decoder, allowing the model to possess the original network's VSR capabilities, and the encoder to extract video semantic features from the video sequence. 
In this stage, we constrain the models with the decoder's CTC loss and CE loss as shown in Equation \ref{stage1_loss}. 
The second stage is the enhancement stage, where we lock the encoder and decoder's parameters and begin training the generator and discriminator. 
Through the adversarial interaction between generator and discriminator, 
the generator gradually learns to transcribe video features into audio semantic features and progressively learns to complete the video semantic features. 
This stage employs adversarial loss between generator and discriminator as constraints and includes a distance loss between video semantic features and audio semantic features as a regularization term in the generator's loss, 
as detailed in Equation \ref{jepkd_loss}. 
The third stage is called the refinement phase, where we lock the parameters of the encoder, generator, and discriminator, 
and only train the decoder to fine-tune it to adapt to the updated input more closely matching the audio semantic features. 
This stage uses the Equation \ref{stage3_loss} as the loss function which omits  $L\left(a,G\left(z,v\right)\right)$ compared to the stage 1.

{\small
\begin{equation}
   \begin{aligned}
   L_{\text {stage } 1}&= \lambda L_{C T C}(x, y)+\gamma L_{L 1}(a, G(v, z)) \\ 
   &+(1-\lambda-\gamma) L_{G E}(x, y)
   \end{aligned}
   \label{stage1_loss}
\end{equation}
}
{\small
\begin{equation}
   \begin{aligned}
   L_{\text {stage } 3}&= \lambda L_{C T C}(x, y)+(1-\lambda) L_{G E}(x, y)
   \end{aligned}
   \label{stage3_loss}
\end{equation}
}

Because GAN models are prone to imbalance during training, 
we designed this three-stage training process to ensure that the input to the generator is a stable and regular vector in feature space, 
which guarantees the stability of the generator during the optimization process. 
In other words, we introduced the JEP-KD for further optimization when the original model reached its theoretical upper limit. 
More details about the models and training processes are discussed in Section \ref{experiment}.

\begin{table}
\caption{Results on the CMLR dataset}
\label{table}
\small
\renewcommand\arraystretch{1.5}
\setlength{\tabcolsep}{3pt}
\begin{tabular}{p{65pt}|p{52pt}|p{38pt}<{\centering}|p{40pt}<{\centering}|p{20pt}<{\centering}}
\hline
Methods&Pre-training Set&Training Set&Total Size (hours)&CER \\ \hline
CSSMCM \cite{cssmcm}&-&CMLR&61&32.48 \\ \hline
LIBS \cite{duibi1}&-&CMLR&61&31.27 \\ \hline
CTCH \cite{ctch}&-&CMLR&61&22.01 \\ \hline
CT-MIR-Net \cite{ctmirnet}&-&CMLR&61&21.45 \\ \hline
Ours&-&CMLR&61&19.92 \\ \hline
Ours+JEP-KD&-&CMLR&61&14.26 \\ \hline
Ours+JEP-KD&CNCVS+Ours&CMLR&361&11.97 \\ \hline
Ma's model \cite{masmodel}&-&CMLR&61&9.10 \\ \hline\hline
WeNet(ASR)&WeNetSpeech&-&10,000+ (Audio)&2.36 \\ \hline
\end{tabular}
\label{table:tab1}
\end{table}

\section{Experiment}
\label{experiment}
\subsection{Dataset}
We conducted preliminary experimental verification on the CMLR dataset \cite{cmlr}\footnote{https://www.vipazoo.cn/CMLR.html}, 
which is a Chinese sentence-level lip-reading dataset. 
It was collected by the Visual Intelligence and Pattern Analysis (VIPA) group of Zhejiang University. 
The CMLR dataset contains 11 speakers, with a total of 102,072 spoken sentences, 
including 71,448 in the training set, 20,418 in the test set, and 10,206 in the validation set.
In the early stage, we referred to Ma's work \cite{masshujuchuli}\footnote{https://github.com/mpc001/auto\_avsr/tree/main/preparation} and preprocessed the dataset. 
Specifically, we extracted the lip-centered lip area of 88×88 size from the video data and saved it as a numpy array. 
It is worth noting that we did not use $randomcrop$ to randomly intercept an 88×88 area from the 96×96 image. 
This was to ensure the speed of data reading during model training, 
reduce the amount of calculation during data reading, and improve the experimental efficiency. 
In addition, we maintain the three-channel input of the RGB image instead of using a single-channel input. 
In terms of speech features, we used a pre-trained WeNet \cite{wenet1, wenet2} speech recognition model\footnote{https://github.com/wenet-e2e/wenet/blob/main/examples/wenetspeech\\/s0/README.md} trained on WeNetSpeech \cite{wenetspeech} dataset for feature extraction, 
and saved the output of the model encoder as a numpy array. 
The reason why we use this model is not only that it can achieve good speech recognition performance on this dataset, 
but also because its encoder structure is consistent with our video encoder structure, and it uses audio input with a sampling rate of 16,000Hz. 
After fbank feature extraction and quadruple length downsampling, the feature length is consistent with the video feature length of 25 fps, 
so that the length and dimension of our video semantic features are the same as the audio semantic features. 
In order to accelerate the model training speed, the pretrained weights of the ASR model are loaded at the start of training.
So we use the vocabulary of the ASR model as the vocabulary of our VSR model. The length of the vocabulary is 5,536, but the actual CMLR dataset only contains about 3,300 different characters. 
 Increasing the vocabulary increases the difficulty of the recognition task to a certain extent, 
 but our experiments mainly focus on performance improvement of the model before and after the introduction of JEP-KD.

\subsection{Evaluation Metric}
Since the smallest unit of the Chinese language is the character, 
similar to most Chinese lip-reading or speech recognition tasks, 
we use CER (Character Error Rate) instead of WER (Word Error Rate) as the model evaluation metric. 
The formula for CER is shown in Equation \ref{cer}, 
where $S$, $D$, $I$ represent the number of substituted, deleted, and inserted characters, respectively, 
required to transform the reference into the hypothesis, and $N$ is the total number of characters in the reference.

{\small
\begin{equation}
   C E R=\frac{S+D+I}{N}
   \label{cer}
\end{equation}
}

\subsection{Training Protocol}
The model structure and loss functions during the training process are as previously described. 
On the CMLR dataset, we set the first phase of training for 20 epochs, the second phase for 10 epochs, and the third phase for 2 epochs. 
All four models use the Adam optimizer, with the encoder and decoder using warmup learning rate scheduling, maximum learning rate of 0.001, and warmup steps of 5,000. 
We set the loss weight $\lambda$ to 0.3, $\gamma$ to 0.1.

Due to the JEP-KD being a universal architecture for lip-reading models, 
the primary objective of our initial experiments is to verify the enhancement of lip-reading models provided by the JEP-KD structure. 
Therefore, we mainly focus on the comparison of model's performance before and after the introduction of the JEP-KD structure. 
To address this, a controlled experiment was conducted, during which 2 models were trained for 32 epochs on the CMLR training set, without employing data augmentation techniques. 
The first model underwent only the warmup stage of training, whereas the second model proceeded through both the second and third stages. 
For the first model, starting only the first phase of training is simply equivalent to introducing the common knowledge distillation structure used in current VSR works.

\subsection{Analysis}
Table \ref{table:tab1} shows that after incorporating the JEP-KD structure, 
the character error rate (CER) of the lip-reading model decreased further, from 19.92\% to 14.26\%. 
Compared to other methods listed in the table, our baseline model did not undergo significant updates. 
It was similar to the latest mainstream methods, which is why the results obtained from training on the same dataset were not much different, with CERs around 20\%. 
However, after adding the JEP-KD structure, our model experienced a more substantial decrease in CER, dropping to 14.26\%, a reduction of about 5 percentage points. 
Considering the already low CER, a 5-percentage-point decrease is quite substantial. 
However, the last row of the Table \ref{table:tab1} indicates that the ASR model could achieve around 2\% CER without being fine-tuned on this dataset, 
suggesting significant information loss during the knowledge distillation process or that the predicted model's output in the feature space still has a considerable distance from the audio features. 

Additionally, we performed pre-training using part of the CNCVS dataset \cite{cncvs} and our own collected dataset to test the performance of the JEP-KD structure under large-quantity conditions. 
Table \ref{table:tab1} indicates that after adding nearly 300 hours of pre-training data, the CER was further reduced to 11.97\%. 
This result demonstrates that the JEP-KD structure, particularly the predictive structure part, maintains a certain level of stability during the training process on large-scale datasets with complex scenes.

\section{Conclusion}
In this paper, we proposed a knowledge distillation framework for lip-reading based on the JEP-KD structure, 
which introduces a generator in the embedding layer to translate video semantic features into audio semantic features. Also, it can predict the regularity loss of video semantic features compared to audio semantic features and complete the semantic features. 
Additionally, we implemented a four-model, three-stage training scheme to ensure the stability of the knowledge distillation of the JEP-KD structure during the training process, enabling the model to achieve optimal training results. 
Experiments demonstrate that the knowledge distillation framework based on the JEP-KD structure can enhance the performance of lip-reading models. 
However, it is regrettable that the experimental results indicate that there is still a significant gap compared to semantic recognition models with similar structures, suggesting that there is substantial room for research in the completion of semantic features. 
Whether we can further narrow this gap by enhancing the predictive model's capabilities remains a question worth exploring. 
On the other hand, verifying the universality of the JEP-KD structure among different lip-reading models, 
as well as employing an AVSR as teacher network combined with the JEP-KD structure to further improve the performance of lip-reading models, is also part of our next research content.

\end{document}